\documentclass[10pt,twocolumn,letterpaper]{article}

\usepackage{cvpr} 
\usepackage{times}
\usepackage{epsfig}
\usepackage{graphicx}
\usepackage{amsmath}
\usepackage{amssymb}
\usepackage{booktabs}
\usepackage{adjustbox}
\usepackage{tablefootnote}
\usepackage{subcaption}

\usepackage{todonotes}
\usepackage{fontawesome5}

\usepackage{balance}

\usepackage[pagebackref=true,breaklinks=true,letterpaper=true,colorlinks,bookmarks=false]{hyperref}

\begin{document}

\title{Identity-Preserving Aging and De-Aging of Faces in the StyleGAN Latent Space}

\author{Luis S. Luevano \quad
Pavel Korshunov \quad
S\'{e}bastien Marcel \\
Idiap Research Institute\\
Rue Marconi 19, 1920 Martigny, CH\\
{\tt\small \{luis.luevano, pavel.korshunov, sebastien.marcel\}@idiap.ch}
}

\maketitle
\thispagestyle{empty}

\begin{abstract}

Face aging or de-aging with generative AI has gained significant attention for its applications in such fields like forensics, security, and media. However, most state of the art methods rely on conditional Generative Adversarial Networks (GANs), Diffusion-based models, or Visual Language Models (VLMs) to age or de-age faces based on predefined age categories and conditioning via loss functions, fine-tuning, or text prompts. The reliance on such conditioning leads to complex training requirements, increased data needs, and challenges in generating consistent results. Additionally, identity preservation is rarely taken into account or evaluated on a single face recognition system without any control or guarantees on whether identity would be preserved in a generated aged/de-aged face. In this paper, we propose to synthesize aged and de-aged faces via editing latent space of StyleGAN2 using a simple support vector modeling of aging/de-aging direction and several feature selection approaches. By using two state-of-the-art face recognition systems, we empirically find the identity preserving subspace within the StyleGAN2 latent space, so that an apparent age of a given face can changed while preserving the identity. We then propose a simple yet practical formula for estimating the limits on aging/de-aging parameters that ensures identity preservation for a given input face. Using our method and estimated parameters we have generated a public dataset of synthetic faces at different ages that can be used for benchmarking cross-age face recognition, age assurance systems, or systems for detection of synthetic images. Our code and dataset are available at the project page \url{https://www.idiap.ch/paper/agesynth/}

\end{abstract}


\section{Introduction}

%

Face image synthesis has seen substantial advancements in recent years, driven by breakthroughs in generative models like Generative Adversarial Networks (GANs)~\cite{GAN-goodfellow, stylegan-original-karras}. Applications for face image synthesis scenarios include face aging and de-aging, strengthening face~\cite{swiftfaceformer} and age~\cite{pavel_mobilenetv2_age_estimation} verification systems against physical~\cite{Korshunov2024pad, Luevano2023Exploring} and digital~\cite{Korshunov2023deepfakes, Luevano2024Assessing} spoofing attacks, mitigating bias of face recognition (FR) systems across different age groups~\cite{korshunov2025bias}, and improving robustness of cross-age face recognition. In particular, these methods can be used to modify a person's face appearance to an arbitrary age while retaining their identity, allowing us to generate younger and older looking versions of the same subject.  State of the art approaches typically employ Conditional GANs \cite{Age-CGAN-2017ICIP} or Diffusion models \cite{latent-diffusion-banerjee-2023} that condition the face transformation on a specific target age group, producing realistic results at the expense of sacrificing controllability, involving complex training procedures, extensive computing resources, and requiring large amounts of labeled data for fine tuning. 
\begin{figure}[!tp]
    \centering
    \includegraphics[width=1\linewidth]{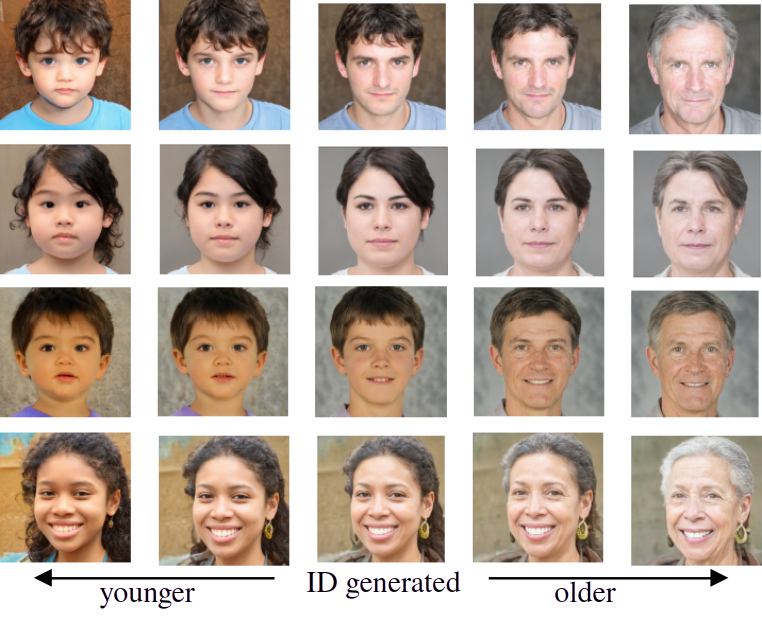}
    \caption{Samples from fully synthetic dataset generated. IDs generated using Langevin sampling \cite{geissbuhler2024syntheticfacedatasetsgeneration} and linear approach for aging and de-aging the synthetic identities.}
    \label{fig:langevin_samples}
\end{figure}

\begin{figure*}[!tp]
    \centering
    \includegraphics[width=0.9\linewidth]{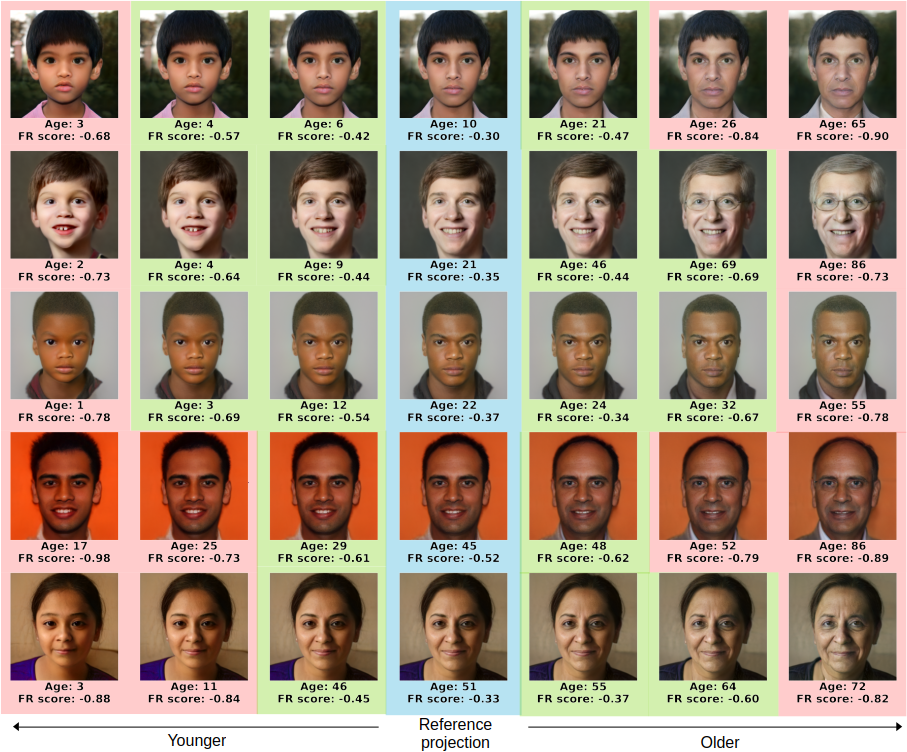}
    \caption{UTKFace sample re-projections from different age groups showing variation in estimated age and FR score from the cosine distance of the EdgeFace embedding to the real-world reference. The column with blue background shows the re-projection without aging/de-aging, the aged/de-aged faces over green are verified against the original, while those over red are rejected. }
    \label{fig:age-samples}
\end{figure*}

When faces are synthetically projected aged or de-aged, it is difficult to preserve identity, which is essential in applications where age transformation must retain enough distinctive features for accurate recognition purposes. While some face aging methods incorporate identity consistency losses~\cite{IPCGAN-2018CVPR} or use pre-trained face recognition models~\cite{AW-GAN-2022NeuralCompAppl} to verify the synthesized face’s identity, most approaches are limited by their dependence on a single face recognition model feature space when conditioning GANs~\cite{Age-CGAN-2017ICIP, FA-CGAN-2022Atkale} and posterior face verification~\cite{IPCGAN-2018CVPR}. This reliance on using existing face recognition pretrained models can potentially introduce biases, as the preservation of identity may vary when evaluated using different recognition systems. As such, it is important to include evaluations using different face recognition systems, whereas current age synthesis methods rarely report face recognition accuracy performance on more than one system. 



In particular, GAN approaches operate in continuous latent spaces and lack predefined steps or directions to represent a person's age, making them challenging to navigate effectively for cross‐age synthesis while preserving identity. In this work, we address this gap by conducting targeted experiments in the StyleGAN2 \cite{stylegan2-original-karras} latent space, a GAN variant popular for its ability to represent style-based attributes, to identify a global ``age synthesis direction'', examining the extent to which identity can be preserved across different age groups. We further enhance identity retention by selecting relevant features, both with and without explicit age or face recognition labels. Furthermore, we generate and release a fully synthetic dataset using face age synthesis approach and synthetic identity sampling, illustrated in Figure \ref{fig:langevin_samples}.  


Concretely, in this work we review the current state of face age synthesis research and propose a simple data‐efficient strategy that operates within the StyleGAN2 $W$ latent space. We explore the challenges of navigating this space while maintaining robust face verification performance (as shown in Figure \ref{fig:age-samples}) using a linear approach for age synthesis and improve identity retention by employing various feature selection mechanisms, including comparisons using reconstructions from Principal Component Analysis (PCA) and Linear Discriminant Analysis (LDA) subspaces given the original projected datasets in the latent space. Unlike conventional methods that rely on conditioning GANs, our approach leverages the readily-available StyleGAN2 structured and rich latent space for straightforward edits that achieve both aging and de‐aging without re-training GAN encoders and decoders. We assess how different feature selection approaches, particularly those emphasizing age and identity components, affect identity preservation when using the selected components in latent space manipulations. We then map the relationship between apparent age, StyleGAN2 scalar step, and assess the identity preservation bounds via curve fitting and root analysis to propose a more practical approach for synthesizing specific ages using this latent space. Finally, we generate a fully synthetic public dataset\footnote{\label{footnote:code}Code and dataset available at the project page:\\ \url{https://www.idiap.ch/paper/agesynth/}} with aged face images by applying the aging approach on synthetically generated identities with Langevin sampling \cite{geissbuhler2024syntheticfacedatasetsgeneration}.

This paper has the following contributions:
\begin{enumerate}
    \item We provide a baseline streamlined and data-efficient approach to face age synthesis using latent editing within StyleGAN2, reducing the need for conditioning networks and extensive labeled datasets by extending linear approaches using Support Vector Regression.
    \item We propose different feature selection strategies and evaluate them using the latents from the StyleGAN space, including reconstruction approaches based on Principal Component Analysis (PCA) and Linear Discriminant Analysis (LDA), that effectively improve identity preservation and apparent age synthesis within the StyleGAN2 latent space.
    \item We propose a set of specific experiments and visualization techniques to analyze the extent of identity preservation within the StyleGAN2 latent space across age modifications, identifying thresholds where identity starts to degrade.
    \item We release a practical toolset\footnotemark[1] for synthesizing faces to a specific age with our findings: linear age direction, weights of $W$ components, fitted curves per age group, and code for finding the corresponding StyleGAN2 scalar step.
    \item We generate and release a synthetically generated dataset\textsuperscript{\ref{footnote:code}} using our approach with $20$k identities and 10 age variations per identity, providing replicability, and addressing privacy concerns over the sensitivity of cross-age data availability.
    
\end{enumerate}

The paper is organized as follows: Section \ref{sec:related-work} describes the related work for face age synthesis methods, Section~\ref{sec:methodology} details approach for efficient controlled latent editing and feature selection, Section~\ref{sec:experimental-setup} shows our experimental setup, in Section \ref{sec:results} we present the experiment results, and we conclude in Section \ref{sec:conclusion}.




\section{Related work}
\label{sec:related-work}

In this section we present related methods on face age synthesis, for Deep Learning-based approaches. It is possible to classify these methods on those based on Translation, Sequences, Conditions, and Style \cite{face-age-synthesis-survey-2022ISCV, face-age-synthesis-survey-2023PattermRecognition}.

\paragraph{Translation-based methods.}
These approaches use GAN methods to age faces by training generators to address consistency among specific age gaps. The authors of F-GAN \cite{FAGAN-2018CVPRW} propose to use CycleGAN \cite{cyclegan-2017ICCV} losses and face age estimators to improve synthesis between smaller gaps in age groups. E2E‑CycleGAN \cite{E2E-CycleGAN-2019} trains using edge maps and generates them before synthesizing the textures, helping preserve geometry. FA‑CGAN \cite{FA-CGAN-2022Atkale} uses the CycleGAN backbone and with additional convolutional blocks, an alternative discriminator, and multi-scale feature fusion to improve face image quality including fine details such as wrinkles. 

These approaches show limitations because they must train multiple generator–discriminators with pairs for every age gap, while relying on cycle and identity losses that depend on face recognition backbones. They also rely on training specific pairs of generators for aging and de-aging predetermined age gaps, becoming computationally expensive to age or de-age outside of these age groups, requiring additional generators.

\paragraph{Sequence-based methods.}
These methods decompose the age gaps into a sequence of incremental aging steps. Recurrent Face Aging (RFA) \cite{Recurrent-face-aging-RFA-2016CVPR} uses optical–flow‑normalized eigenfaces as input and trains Gated Recurrent Unit (GRUs) to model long and short-term dynamics of aging,  transferring high‑frequency details from each group's nearest neighbors. Triple‑GAN \cite{triple-GAN-2020CVPRW} aligns three simultaneous translation paths with a “triple” loss so that the generated faces along the different paths converge to the same target age. PFA‑GAN \cite{PFA-GAN-2021TIFS} divides the generator itself into gated sub‑networks, each tuned to a narrow age span, yielding smoother trajectories and mitigating ghosting artifacts.

Shortcomings of these approaches include cases when we attempt to synthesize over large age gaps, such as error accumulation and an increase in complexity due to the chain of generators. Furthermore, it is not possible to synthesize other age gaps different from those contemplated at training time.

\paragraph{Conditional-based methods.}
Conditional GANs use the desired age label as input for the generator and discriminator. Age‑CGAN \cite{Age-CGAN-2017ICIP} searches the latent space with identity‑preserving optimisation so that FaceNet embeddings stay close while the age code changes. IPCGAN \cite{IPCGAN-2018CVPR} adds perceptual, age‑classification, and identity preservation losses at training time. Other methods, such as ChildFace \cite{ChildFace-2020BIOSIG}, AW‑GAN \cite{AW-GAN-2022NeuralCompAppl}, and AgeTransGAN \cite{AgeTransGAN-2022-ECCV} strengthen detail using coarse‑to‑fine generators, wavelet‑domain critics, and multitask discriminators, respectively. In ChildGAN \cite{ChildGAN}, the authors train separate StyleGAN2 instances per gender and perform logistic regression in the latent space for attribute augmentation.

These methods show limitations in their dependence on explicit target age labels at train time and test time. They also need large and accurately annotated age datasets and employ complex conditioning pipelines for synthesizing multiple attributes such as age, gender, or pose. Their identity preservation mechanisms also rely heavily on a single recognition model embedded in the loss, potentially limiting the usability of the output to a particular face recognition backbone.

\paragraph{Style-based methods.}
These approaches shift the focus from re‑training to navigating an already expressive latent space. SAM \cite{SAM-Style-2021ACM} places a head for age regression on top of StyleGAN2 and back propagates the regression gradient to the $W$ projection modules, regularized by identity and cycle constraints. FAM \cite{Children-Deep-Feature-Aging-FAM-2020ICPR} directly manipulates FaceNet features before re‑synthesising the image. Other methods include IricGAN \cite{IricGAN-face-editing-2023Ning}, which uses multi‑scale attribute regression, and DyStyle \cite{DyStyle-2023WACV}, comprised of dynamic expert modules selected per attribute. HDA‑SynChildFaces \cite{Falkenberg-HDASynChildFaces-Frontiers-2024} employs multiple SVM hyper‑planes in the latent space to generate a large dataset focused on children faces only. Diffusion models for aging faces, such as \cite{latent-diffusion-banerjee-2023}, use text prompts and labels as input to the method, delegating the text-to-image translation task to cross-modal alignment methods, thus sacrificing controllability.

Even though these approaches are based on latent editing, the training process is often made more complex by re-training the generator, adding extra networks at training time, introducing complex training losses, and finding latent directions using \emph{multiple} SVMs. Furthermore, these approaches do not attempt to explore which latent components meaningfully affect age or identity, which may also introduce artifacts and undesired identity changes. In this paper, we address these shortcomings by using the readily-available StyleGAN2 generator without re-training or training additional convolutional modules, finding a \emph{single} latent direction with little amount of training data, and improving identity preservation by selecting relevant latent vector components to age and identity.

\section{Editing in the StyleGAN latent space}
\label{sec:methodology}

In this section, we will detail the method for latent analysis, latent edits, and feature selection.

\subsection{Latent analysis and linear latent editing} 
\label{subsec:latent_analisis_edit}

The latent analysis approach is an extension of the method from \cite{laurent-latent-edit-ijcb-2021}, which begins by projecting real face images from the image space into the $W$ $512\text{-}d$ dimensional latent space of StyleGAN2, where the $w$ latent vectors are used for age attribute analysis and editing. We extend the approach in \cite{laurent-latent-edit-ijcb-2021} for finding the direction in the latent space from `young' to `old'. 
Our approach uses a Linear SVR to fit latents in the StyleGAN2 $W$ space to their original age labels $y\in\mathbb{R}$. Linear SVR fits the hyperplane containing most of the samples inside its surrounding margin. This hyperplane is defined by $\{y= \theta^{T}\tilde{w} | \theta, \tilde{w} \in \mathbb{R}^{|W|+1}\}$, where $\theta$ contains the hyperplane parameters (bias $b$ and coefficients $\lambda$) and $\tilde{w}$ a latent vector $[1 \text{  }w]$ in $R^{|W|+1}$. Given that we have a known latent $w_0$ with age value $y_0$, we want to find a new latent $w_1$ reflecting the desired change in age (we want to age or de-age $w_0$). By projecting the vector $\theta$, used to predict the $y$ age value, onto the latent vector $\tilde{w}_0$, we can find the direction in which the age of $w_0$ changes within $W$ space. We can formulate $\tilde{w_1}=\tilde{w_0}+c\tilde{w}_0$, with $c$ given by $\text{Proj}_{\tilde{w_0}}\theta = \frac{\theta\cdot \tilde{w_0}}{\tilde{w_0} \cdot \tilde{w_0}}$. After substitution, we have $\tilde{w_1}$ as :
\[
\tilde{w_1}=\tilde{w_0}+\theta
\]
\[
[1\text{  }w_1]=[1\text{  }w_0] + [b\text{  }\lambda]
\]

With $\lambda$ corresponding to the vector changing the components of $w_0$ in the $W$ space. Finding the direction of $\lambda$ as $\hat{\lambda}=\frac{\lambda}{||\lambda||_2}$, a normalized unit vector, allows us to move $w_0$ to the desired $w_1$ by scaling $\hat{\lambda}$ using scalar multiplication $s$ as:

\begin{equation}
\label{eqn:linear_scalar}
w_1 = w_0 + s\mathbf{\hat\lambda}
\end{equation}

By projecting $w_1$ into pixel space, we obtain a face of the person appearing older or younger depending on the sign of $s$, which its value represents the number of years. What remains to be found is the mapping between the values of scalar $s$ and the number of years by which we want to age/de-age the person. We are interested in mapping the changes in age in function of $s$ because obtaining $w_1$ from $w_0$, the $y$ value in years, and the age direction $\hat\lambda$ depends on moving through a linear hyperplane, whereas the latents may not correspond linearly to their $y$ age value and their placement in the latent space may be depending on other factors, such as the apparent age of the original face.

\subsection{Identity preservation in the latent space}
\label{sec:idpreserved}

\begin{figure}
    \centering
    \includegraphics[width=1.05\linewidth]{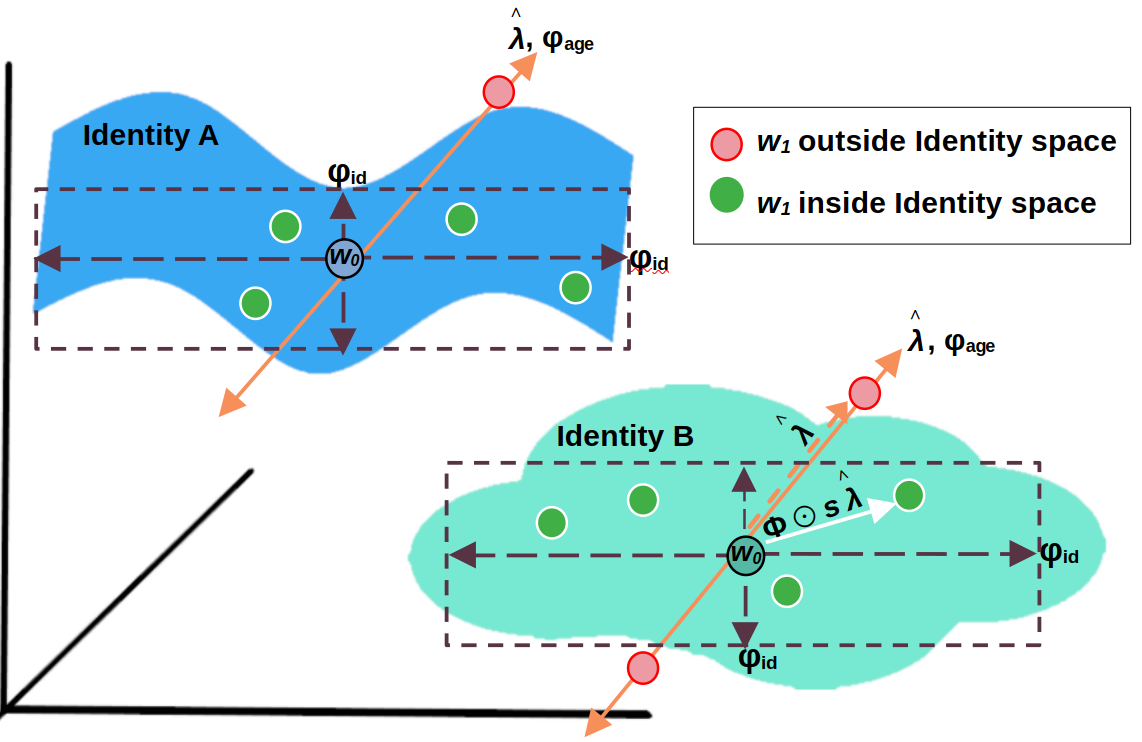}
    \caption{Controlled latent edit approach with identity preservation in the $W$ space. We age and de-age each sample projection $w$ using the scaled age direction $\mathbf{s\hat\lambda}$, controlled by the rate weight $\Phi$, which includes the change in the selected latent vector components relevant to identity ($\varphi_{\text{id}})$ and age ($\varphi_{\text{age}}$). This re-scales the change in the age direction, aiding in synthesizing points inside the identity region of each subject inside the latent space.}
    \label{fig:controlled_latent_edit}
\end{figure}

When using Eq.~\ref{eqn:linear_scalar}, we move linearly using all the components of the StyleGAN2 vector in the latent space indiscriminately by multiplying them with the floating-point scalar step $s$. We refer to the approach described by Eq.~\ref{eqn:linear_scalar} as a linear baseline approach, since it does not involve any identity constraints.

We can extend the linear baseline approach 
by multiplying only a selected subset of the  components in the StyleGAN2 vector, with the aim to find those components that contribute the most to the identity and age changes. Hence, we can modify Eq. \ref{eqn:linear_scalar} to include a $\Phi$ term describing an additional weight vector regulating the change per each component of the scaled $\mathbf{\hat\lambda}$ age direction vector as follows:

\begin{equation}
\label{eqn:linear_scalar_weight}
w_{1} = w_0 + \Phi \odot s\mathbf{\hat{\lambda}}
\end{equation}

The $\Phi$ vector re-scales and re-shapes the region followed by the synthesized latent vector $w_1$ through element-wise multiplication ($\odot$), constraining the magnitude of the directions for different components of the latent vector. We compute this $\Phi$ vector from selecting features relevant to age ($\varphi_{\text{age}}$) and identity ($\varphi_{\text{id}}$) from the latent components.
In the linear baseline case, this ${\Phi}$ scaling vector represents an equal change in all the components from $\mathbf{\hat\lambda}$ and $w$. Figure \ref{fig:controlled_latent_edit} illustrates this approach.

\paragraph{Feature selection.}
\label{sec:featanalysis}

The main goal of feature selection in this work is to compute $\Phi$ from Eq. \ref{eqn:linear_scalar_weight} to regulate the change in our latent vector components. We propose multiple ways of computing $\Phi$. In general, we employ PCA and LDA to assess component importance through sub-space projection and reconstruction. For this purpose, we use two sets of \emph{standardized} latent vectors: one from a dataset with identity labels ($V_{\text{id}}^{n\times |W|}$) (for PCA and LDA) and another one from a dataset with age group labels ($V_{\text{age}}^{m\times |W|}$), (only for LDA).

\paragraph{Computing $\Phi$ using PCA.}
We use the $V_{\text{id}}$ set to compute {$\Phi_{\text{PCA}}=\varphi_\text{id-PCA}$}. Firstly, we compute the eigen-decomposition for $aP_\text{PCA} = \text{Eig}(V_{\text{id}}V_{\text{id}}^T)$, with $a$ denoting the vector of unsorted eigenvalues of the projection basis $P_\text{PCA}$. We calculate the cumulative variance of the largest eigenvalues and select the component indices corresponding to 95\% of variance ($\sigma_\text{PCA}$). We use the selected components to directly set the values of the $\varphi_{\text{id-PCA}}$ mask, since we found that the most relevant components in the PCA basis are the same as in the latent space after projection, reconstruction, and comparison. Consequently, we formulate:
\[
\varphi_{\text{id-PCA}} = \{\rho_i ... \rho_{|W|} | \rho = 1 \text{ if } a_i \in \sigma_\text{PCA} >95\%, 0 \text{ otherwise} \}
\]

\paragraph{Computing $\Phi$ using LDA.}

We find the components important for identity $\varphi_{\text{id-LDA}}$ using $V_{\text{id}}$ with identity labels and those important for age as $\varphi_{\text{age-LDA}}$ with $V_{\text{age}}$ with age group labels.

We compute separate LDA bases, $aP_\text{LDA}$, from the eigen-decomposition of the product of the scatter matrices for each set of latents. The magnitude of the eigenvalues in $a$ represent the ratio of discriminability per class and component, which we use to find the most important discriminative components in the basis, leaving those contributing to 95\% discriminability intact. We set the least important components in $P_\text{LDA}$ as zero vectors, and use this matrix to project the latents into the sub-space. Using the matrix pseudoinverse $P_\text{LDA}^{+}$, we reconstruct the projected latents from the LDA space back to the latent space to compare them against the original latents. We denote this reconstruction $V*$.

Each pair of standardized latents $V$ and reconstructed latents $V*$ is used to calculate three different distance metrics $\Psi(V,V*)$: Mean Squared Error (MSE), 1-D Wasserstein distance, and Covariance. We compute the mean per component of $\Psi(V,V*)$ as $\psi^{|W|}$. We calculate the mean $\mu_\psi$ as threshold for feature selection.
We formulate $\varphi_{\text{LDA}}$ when using MSE and 1-D Wasserstein as:
\[
\varphi_{\text{LDA}} = \{\rho_i ... \rho_{|W|} | \rho = 1 \text{ if }  \psi_i < \mu_\psi, 0 \text{ otherwise} \}
\]
and for Covariance as:
\[
\varphi_{\text{LDA}} = \{\rho_i ... \rho_{|W|} | \rho = 1 \text{ if }  \psi_i > \mu_\psi, 0 \text{ otherwise} \}
\]

We use these formulations to calculate the masks $\varphi_\text{id*-LDA}$ and $\varphi_\text{age*-LDA}$ corresponding to each different distance metric. These masks correspond to \emph{all} the components important to identity or age. Hence, we find an overlap of the components relevant to both identity and age as:
\[\varphi_{\text{id}\land \text{age}} = \varphi_{\text{id*-LDA}}\land \varphi_{\text{age*-LDA}}
\]
while the components relevant \emph{only} to age and identity are formulated as 
\[
\varphi_\text{age-LDA} = \varphi_{\text{age*-LDA}} \land \neg \varphi_\text{*id-LDA}\]
\[
\varphi_\text{id-LDA} = \varphi_{\text{id*-LDA}} \land \neg \varphi_\text{*age-LDA}
\]
Therefore, we can transform the weighting vector $\Phi$ from Eq.~\eqref{eqn:linear_scalar_weight} by including the features selected using LDA as follows:

\begin{equation} 
\label{eqn:phi_importance} 
    \Phi_{\text{LDA}} = \alpha\varphi_{\text{age-LDA}} + \beta\varphi_{\text{id}\land\text{age}}
\end{equation} 

where the hyperparameters $\alpha$ and $\beta$ are scalars emphasizing the specific components that are relevant only for age and those impacting age and identity at the same time.

When using LDA with the age-labeled dataset, we could have few age group labels, which could potentially limit the effect of LDA since it is more suitable for data with larger number of classes. In our experiments, we therefore applied LDA with larger number of age classes but did not observe any impact in the component selection results.

\subsection{Estimating scalar steps for specific ages}

When we follow Eq.~\eqref{eqn:linear_scalar_weight} for aging and de-aging, we compute StyleGAN latents and the corresponding images corresponding to a discrete small number of scalars $s$ for each age group. For practical application, we need to know scalar values for every desired age $y_r$ we want to generate. 

We then fit polynomial models $p_r(s)$, to approximate $p_r(s)\approx y_r$ based on the latents we pre-computed for the limited number of scalar $s$. When transforming a person from an \emph{original} age $y_{\text{original}}$ to a \emph{desired} age $y_{\text{desired}}$, we select the polynomial curve $p_r$ corresponding to the sample's age range $r$ from $y_{\text{original}}$. To find the original and desired ages in the polynomial curve, $s_{\text{original}}$ and $s_{\text{desired}}$ respectively, we solve for the scalar shifts by computing the roots of ${p_r(s) \;-\; y_{\text{original}} \;=\; 0}$ and 
${p_r(s) \;-\; y_{\text{desired}} \;=\; 0}$.
The solutions correspond to the scalar value at that age, \(s_{\text{original}}\) and \(s_{\text{desired}}\). We compute the net scalar offset as ${\Delta s \;=\; s_{\text{desired}} - s_{\text{original}}}$.
In practice, we only keep physically meaningful solutions by filtering out those outside a functional scalar range. If a real and unique valid solution does not exist for computing the scalar using the polynomial curve in a category, we can fall back to simpler linear fits, separately for the aging and de-aging cases.

\section{Experimental setup}
\label{sec:experimental-setup}
In this section, we describe our experiments in terms of datasets, practical latent editing, features selection, and evaluation pipeline.

\subsection{Datasets}
\label{subsec:data-preparation}

In the experiments, we use UTKFace \cite{UTKFace-2017CVPR} for the age analysis and Color FERET \cite {color-feret} for the identity analysis. 

\paragraph{UTKFace.} This dataset \cite{UTKFace-2017CVPR} contains samples with age data ranging from 0 to 116 years old, with over 20K faces annotated with age, gender, and ethnicity attributes. We use only a small portion of images to simulate data-constrained scenarios. Firstly, we pre-process the images from the UTKFace dataset using CodeFormer \cite{codeformer-super-resolution-neurips-2022} to scale them up to $2\times$ their original resolution, cropped and aligned to match the original StyleGAN2 FFHQ images to a resolution of $1,024 \times 1,024$ pixels. Next, we take a section of the upscaled dataset, dividing it into a training set partition and a testing set partition. The training set consists of $1,336$ samples, where half are younger than 18, and the other half are over $25$. The test set contains $3,494$ samples. We ensure no overlap with the training sets using an age estimation model~\cite{pavel_mobilenetv2_age_estimation}. We also group the samples in each set using a different number of age groups, which is useful depending on the experiment. For $9$ age groups, we divide them for $< 8$, $[8-13)$, $[13,18)$, $[18-25)$, $[25-35)$, $[35-45)$, $[45-55)$, $[55-65)$, and $>65$. When using $4$ age groups, we label the ages for children ($<18$, 312 samples), young adults ($[18-35)$, 1453 samples), middle aged ($[35-65)$, 1284 samples), and senior ($>65$, 286 samples).

\paragraph{Color FERET.} This dataset \cite{color-feret} contains a total of 11K images taken from a semi-controlled environment, spanning $994$ identities, and annotated with gender and race information. 
We filtered the images containing hard-profile poses (close to $90^{\circ}$ in either direction) and processed them using the StyleGAN2 FFHQ face cropper to a $1024 \times 1024$ resolution. The final set used in this paper contains $994$ identities and has $7.95K$ images, averaging $8$ images per subject, only used for training PCA and LDA.

\subsection{Latent editing and feature selection} 
\label{subsec:impact_baseline_scalar}

We add a small scalar $s$ steps along the age direction in $W$ (Eq.\ref{eqn:linear_scalar}), regenerate each image, and estimate its apparent age with MobileNetV2. Using the UTKFace test set split into 9 age groups, we compute the mean estimated age per group for each $s$, obtaining a discrete mapping $s \rightarrow \text{age}$. We then fit linear and polynomial curves per group to get a continuous relationship, which lets us pick the scalar $s$ needed to reach any desired target age (older or younger).

We use two pre-trained face recognition models scoring over 95\% accuracy in cross-age benchmarks, IResNet50~\cite{iresnet50} and EdgeFace-S~\cite{edgeface}, to estimate identity preservation and MobileNetV2 age estimator model~\cite{pavel_mobilenetv2_age_estimation} to evaluate apparent age of the face images. We find verification threshold for both face recognition systems using the standard benchmark IJB-C \cite{IJBC} dataset. We use face recognition and age estimator to evaluate the effect of using linear features (from Linear SVR) compared to feature selection approaches aimed to preserve identity. We use UTKFace testing partition to compute the number of correctly verified age-manipulated samples and their age difference using the estimated age against the sample's original age label. With this difference, we calculate an \emph{age gain} range, using its mean and one standard deviation, representing our confidence interval. We calculate the age gain for the aging/de-aging cases, depending on if we move the latent using a positive scalar to age the resulting face image or in the negative direction to de-age it. In the feature selection approaches, we set $\alpha$ from Eq. \ref{eqn:phi_importance} to $1$. We varied $\beta$ from $0$ (no change) to $0.25$, $0.50$, $0.75$, and $1$ to find the value that leads to the largest age gain preserving the identity. To that end, we compute the \emph{age gain} for the four age group labels as described in Section \ref{subsec:data-preparation}. We interpolate the \emph{age gain} at each rate of correctly verified samples from the test set of UTKFace, allowing us to examine the \emph{age gain} range behavior at different face verification percentages. 

\begin{figure}[!t]
	\centering
	\includegraphics[width=1.0\linewidth]{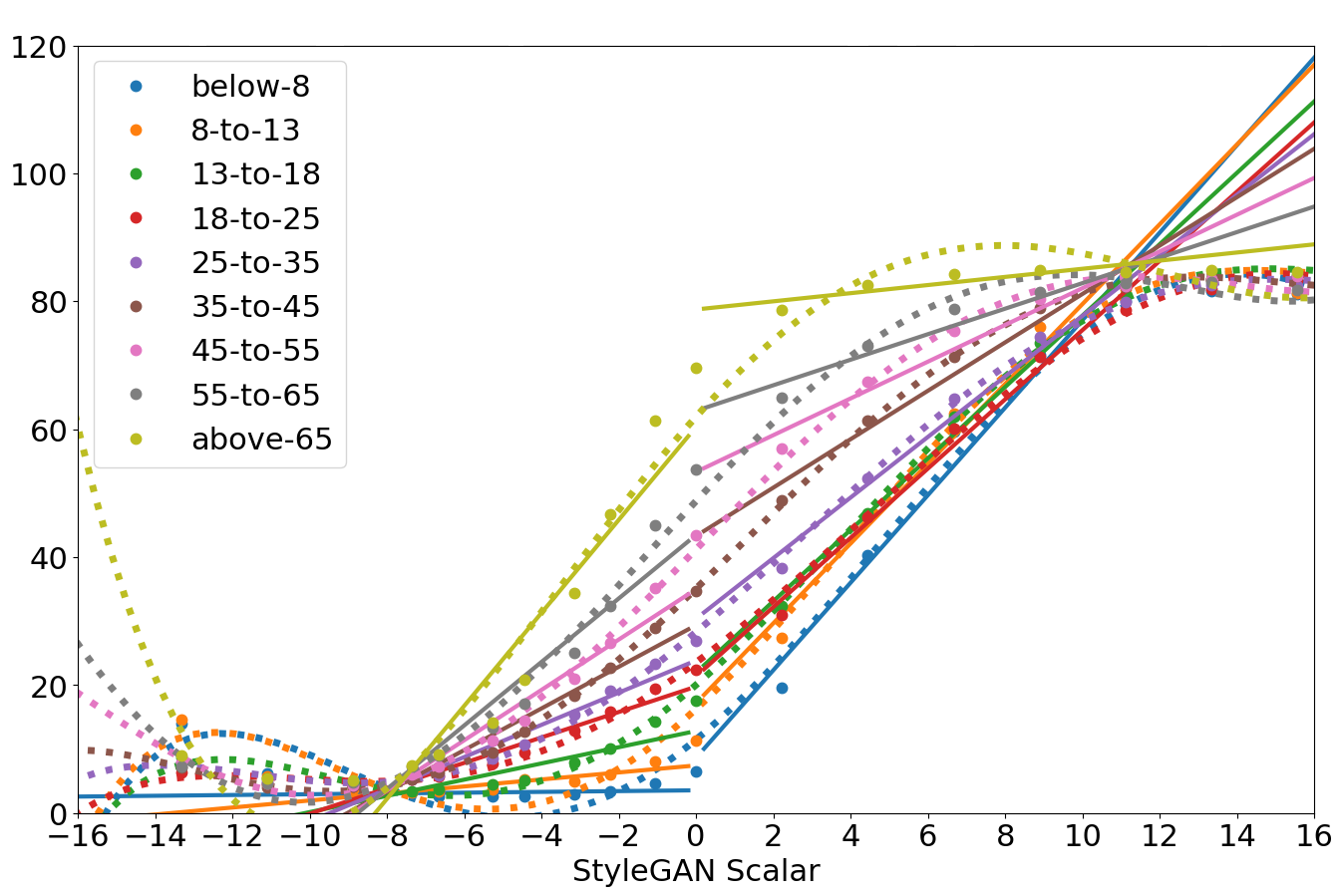}
	\caption{Baseline with linear and polynomial fitting for apparent age change across different age groups compared to the StyleGAN2 scalar change, using UTKFace test set. The scalar $0$ denotes the original re-projected image. We note the differences in impact depending on the subject's age group.}
	\label{fig:utkface-apparent-age}
\end{figure}

\section{Results}
\label{sec:results}

Using real faces with age labels from UTKFace~\cite{UTKFace-2017CVPR}, we apply StyleGAN2 latent space editing approach (see Section~\ref{subsec:latent_analisis_edit} to compute the parameters for identity preserving aging and de-aging.
Then, we compare different feature selection strategies presented in Section~\ref{sec:featanalysis}. Finally, we generate fully synthetic dataset of identities using Langevin sampling~\cite{geissbuhler2024syntheticfacedatasetsgeneration} and aged/de-aged with our approach.




\subsection{Apparent age estimation}

\begin{figure}
	\centering
     \subfloat[Scores from IResNet50 \label{subfig:min-max-apparent-age-real-age-iresnet50}]{%
         \centering
         \includegraphics[width=0.8\linewidth]{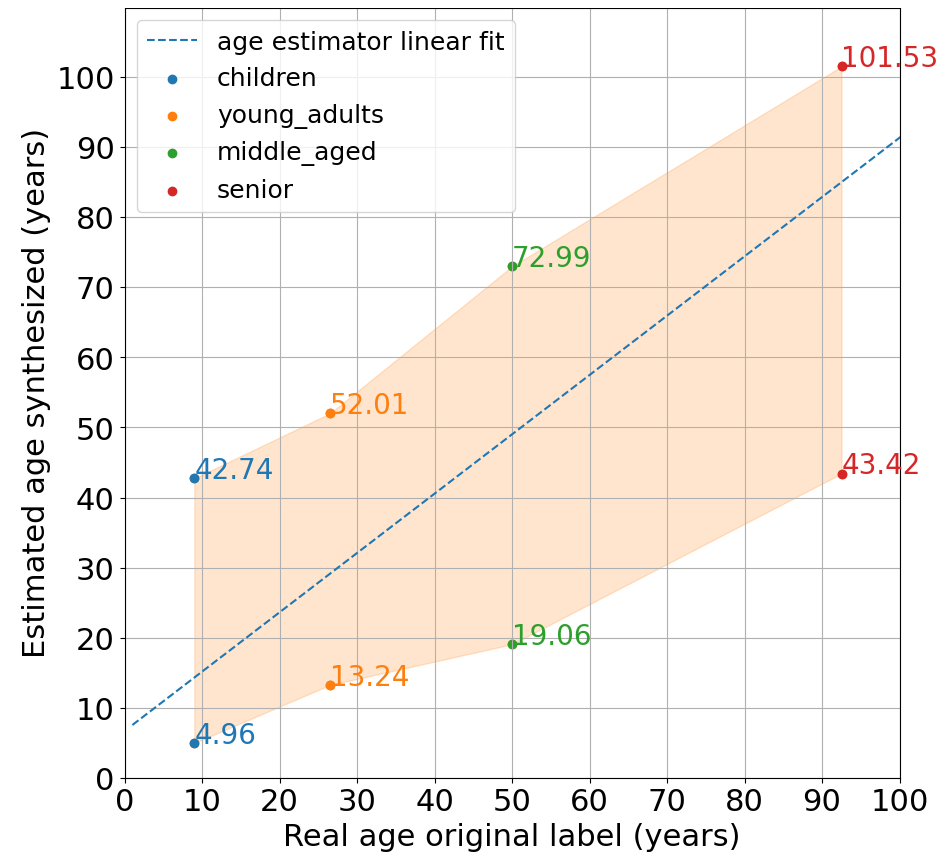}}
     \hfill
      \subfloat[Scores from EdgeFace-S \label{subfig:min-max-apparent-age-real-age-edgeface}]{%
        \centering
	\includegraphics[width=0.8\linewidth]{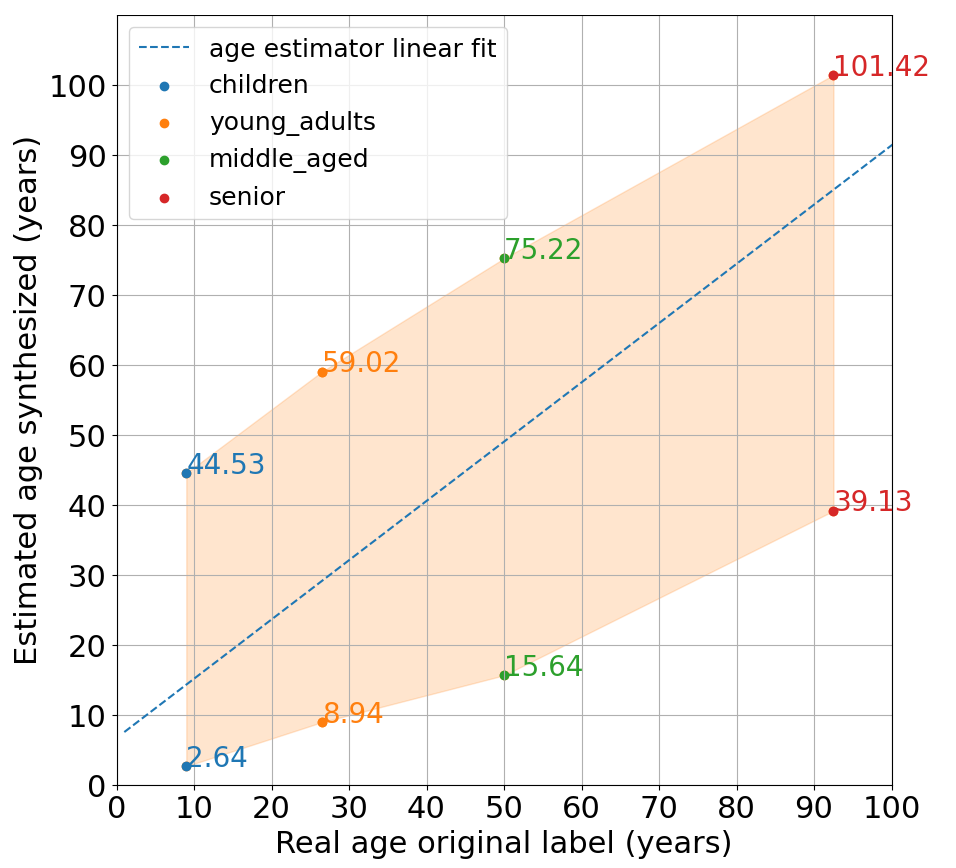}
    }
    \caption{Range of estimated age from synthesized images, with subjects from distinct age groups while preserving identity on 75\% of samples in UTKFace test set. We note the extended range of values in years with EdgeFace-S when compared to IResNet50, in both aging and de-aging cases.}
    \label{fig:estimated_age_range}
\end{figure}

\begin{figure*}[!hptb]
    \centering
    \includegraphics[width=1.0\linewidth]{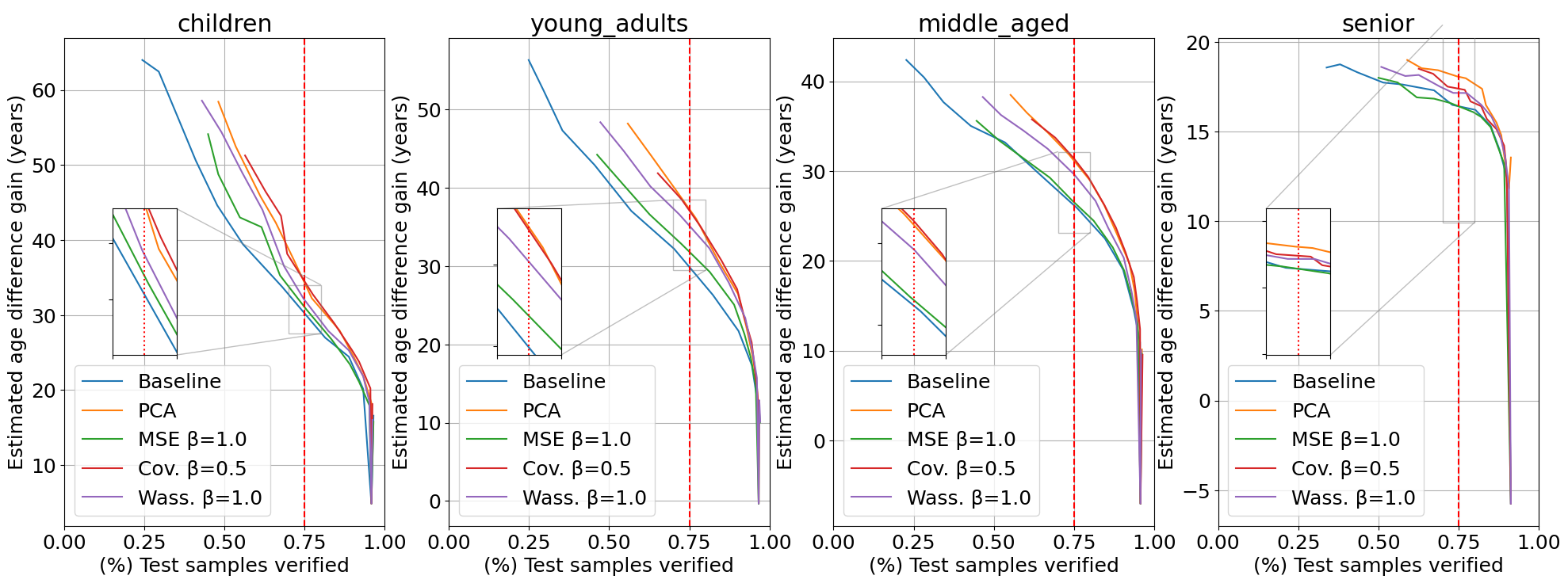}
    
    \caption{Age gain per number of samples verified when aging faces. We show face verification performance against age gain across different age categories. We highlight the differences at 75\% of verified samples from the test set while using EdgeFace-S as face recognition system. The metrics for MSE, Covariance, and Wasserstein distance are computed using LDA bases for projection and re-construction.}
    \label{fig:age_categories_average_verification_edgeface}
\end{figure*}

Figure~\ref{fig:utkface-apparent-age} illustrates the change in the apparent age for different scalar $s$ from Eq.~\ref{eqn:linear_scalar} computed separately for each of the 9 age groups (see Section \ref{subsec:impact_baseline_scalar} for details). 
The scalar value $0$ corresponds to the re-projected original image, the negative scalar values correspond to de-aged images, and the positive scalar values correspond to the aged images. Figure~\ref{fig:utkface-apparent-age} shows that  the re-projection of the original faces (scalar value $0$) appear younger as per the MobileNetV2 age estimator~\cite{pavel_mobilenetv2_age_estimation}. We notice non-linear behaviors when the face is aged or de-aged, especially, if the original age of the subject in the older or younger age groups. The figure also shows polynomial curves fitted to the data, which we can use to compute more accurate step-by-step increments when manipulating the age in the latent space.
\begin{figure}[!b]
    \centering
    \includegraphics[width=1\linewidth]{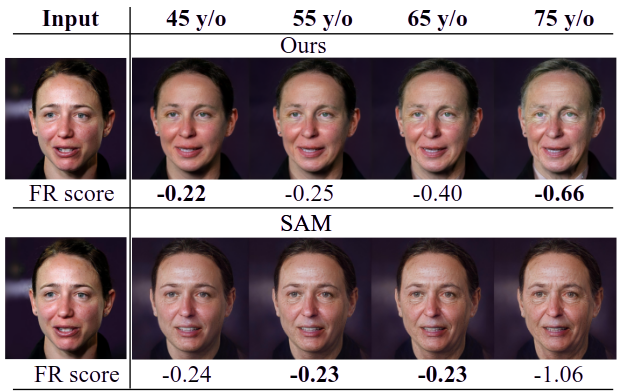}
    \caption{Age synthesis to specific target ages of our approach and SAM \cite{SAM-Style-2021ACM} The face recognition (FR) score is the cosine distance of each image to the input using EdgeFace-S embeddings.}
    \label{fig:comparison-sam}
\end{figure}

\subsection{Maximum aging ranges while preserving ID}


We age/de‑age faces with the baseline method and check identity with two face recognition models at cutoff rate of $75$\% of samples. We then linearly fit the estimated age of the original samples and hence find the age‑gain for each age group. This age-gain is shown in
Figure \ref{subfig:min-max-apparent-age-real-age-iresnet50} as the highlighted region, within which  aged/de-aged faces are matched with the original images using IResNet50~\cite{iresnet50}, and in Figure \ref{subfig:min-max-apparent-age-real-age-edgeface} using EdgeFace-S~\cite{edgeface}. 

Figure \ref{subfig:min-max-apparent-age-real-age-iresnet50} of IResNet50 shows an obvious trend that the linear baseline approach can age children and de-age seniors by many years, while it does not work well in reverse.
Also, middle age catergories can be both ageed and de-ageed by a large margin.
Figure \ref{subfig:min-max-apparent-age-real-age-edgeface} for EdgeFace-S shows that the apparent age range limits are wider acriss all categories.
For example, EdgeFace can recognize children faces de-aged to 2 years and aged to 30 years, showing the impact of the chosen face recognition system. 

\subsection{Estimated age gain on feature selection methods}

Figure \ref{fig:age_categories_average_verification_edgeface} shows  the performance of linear baseline approach against different feature selection mechanisms as described in Section \ref{sec:featanalysis}. The scores from EdgeFace-S are sued as reference. The figure shows an expected drop in the rate of correctly recognized samples as we move in the StyleGAN2 space, with respect to its apparent age (estimated) and allow us to compare ID-preservation performance of different feature selection strategies. Zoomed boxes emphasize the performance at $75$\% of recognized samples.
All the feature selection methods improve over the baseline in terms of apparent age and the rate of samples correctly verified. In particular, PCA and LDA using Covariance show the most improvements over the linear baseline, with the latter being slightly better. These two approaches perform similarly, with covariance performing slightly better over PCA in the children age group, and PCA being more robust in the senior age group.
Figure \ref{fig:comparison-sam} compares our method and SAM \cite{SAM-Style-2021ACM}, showing better performance qualitatively and quantitatively across larger age gaps.



\subsection{Fully synthetic aged dataset}

We publicly release a $20$k-ID synthetic dataset\footnotemark[1] with $10$ age versions (Figure~\ref{fig:langevin_samples}), generated using Langevin sampling \cite{geissbuhler2024syntheticfacedatasetsgeneration} and our SVR-based latent edit approach from Section~\ref{subsec:latent_analisis_edit}.

\section{Conclusion}
\label{sec:conclusion}


We showed that a simple SVR-based latent direction and feature selection allows identity-preserving age edits in StyleGAN2 without re-training generators. Polynomial scalar–age mappings provide practical target-age control, and PCA/LDA masks improve the range of ID-preserving age edits. Future work includes alternative age estimators and latent navigation strategies.

\section*{Acknowledgement}
This work was funded by InnoSuisse 108.708 INT-ICT.



%



\balance
{\small
\bibliographystyle{ieee}
\bibliography{refs}
}

\end{document}